\documentclass[letterpaper]{article} 
\usepackage{aaai25}  
\usepackage{times}  
\usepackage{helvet}  
\usepackage{courier}  
\usepackage[hyphens]{url}  
\usepackage{graphicx} 
\usepackage{natbib}  
\usepackage{caption} 
\usepackage{algorithm}
\usepackage{algorithmic}
\usepackage{booktabs} 
\usepackage{graphicx} 
\usepackage{float}    
\usepackage{multirow} 
\usepackage{graphicx} 
\usepackage{makecell} 
\usepackage{amssymb}
\usepackage{amsmath}

\usepackage{newfloat}
\usepackage{listings}
\DeclareCaptionStyle{ruled}{labelfont=normalfont,labelsep=colon,strut=off} 
\floatstyle{ruled}
\newfloat{listing}{tb}{lst}{}
\floatname{listing}{Listing}
\title{Adaptive Few-shot Prompting for Machine Translation with Pre-trained Language Models}
\author{
Lei Tang\textsuperscript{\rm 1}, Jinghui Qin\textsuperscript{\rm 1}\thanks{Corresponding author}, Wenxuan Ye\textsuperscript{\rm 2}, Hao Tan\textsuperscript{\rm 1}, Zhijing Yang\textsuperscript{\rm 1}
}
\affiliations{
    \textsuperscript{\rm 1}Guangdong University of Technology  \\
    \textsuperscript{\rm 2}The Chinese University of Hong Kong

    tangtang302958@163.com, scape1989@gmail.com, nbvincentelite@gmail.com, \\ tanhao4869@gmail.com, yzhj@gdut.edu.cn

}

\usepackage{bibentry}

\begin{document}

\maketitle

\begin{abstract}
Recently, Large language models (LLMs) with in-context learning have demonstrated remarkable potential in handling neural machine translation. However, existing evidence shows that LLMs are prompt-sensitive and it is sub-optimal to apply the fixed prompt to any input for downstream machine translation tasks. To address this issue, we propose an adaptive few-shot prompting (AFSP) framework to automatically select suitable translation demonstrations for various source input sentences to further elicit the translation capability of an LLM for better machine translation. First, we build a translation demonstration retrieval module based on LLM's embedding to retrieve top-k semantic-similar translation demonstrations from aligned parallel translation corpus. Rather than using other embedding models for semantic demonstration retrieval, we build a hybrid demonstration retrieval module based on the embedding layer of the deployed LLM to build better input representation for retrieving more semantic-related translation demonstrations. Then, to ensure better semantic consistency between source inputs and target outputs, we force the deployed LLM itself to generate multiple output candidates in the target language with the help of translation demonstrations and rerank these candidates. Besides, to better evaluate the effectiveness of our AFSP framework on the latest language and extend the research boundary of neural machine translation, we construct a high-quality diplomatic Chinese-English parallel dataset that consists of 5,528 parallel Chinese-English sentences. Finally, extensive experiments on the proposed diplomatic Chinese-English parallel dataset and the United Nations Parallel Corpus (Chinese-English part) show the effectiveness and superiority of our proposed AFSP.
\end{abstract}

%

\section{Introduction}
Neural Machine Translation (NMT)~\cite{bahdanau2015neural}, the core of which lies in the encoder-decoder architecture, aims to translate texts in the source language into the target language automatically. NMT is a challenging task since it involves translating text among different languages and requires semantic alignment between languages~\cite{fan2021beyond, costa2022no, yuan-etal-2023-lego}. Even so, it has made remarkable progress in recent years, especially with the emergence of large language models (LLMs) like ChatGPT \& GPT-4~\cite{ouyang2022training}, GLM~\cite{du2022glm}, Llama~\cite{touvron2023llama, dubey2024llama}, etc. Benefiting from the increasing scale of parameters and training
corpus, these LLMs have gained a universal ability to handle various NLP tasks via in-context learning (ICL)~\cite{brown2020language} or prompt engineering~\cite{chen2023unleashing}, which is the process of structuring input text with exemplars and human-written instructions for LLMs, rather than conducting costly task-specific fine-tuning. Unsurprisingly, LLMs with ICL or prompting techniques have shown outstanding potential in machine translation~\cite{zhang2023prompting, zhu2024multilingual,zhang-etal-2023-machine} by constructing elaborate instruction or prompts with different prompting strategies.

\begin{figure}[t]
\centering
\includegraphics[width=0.99\linewidth]{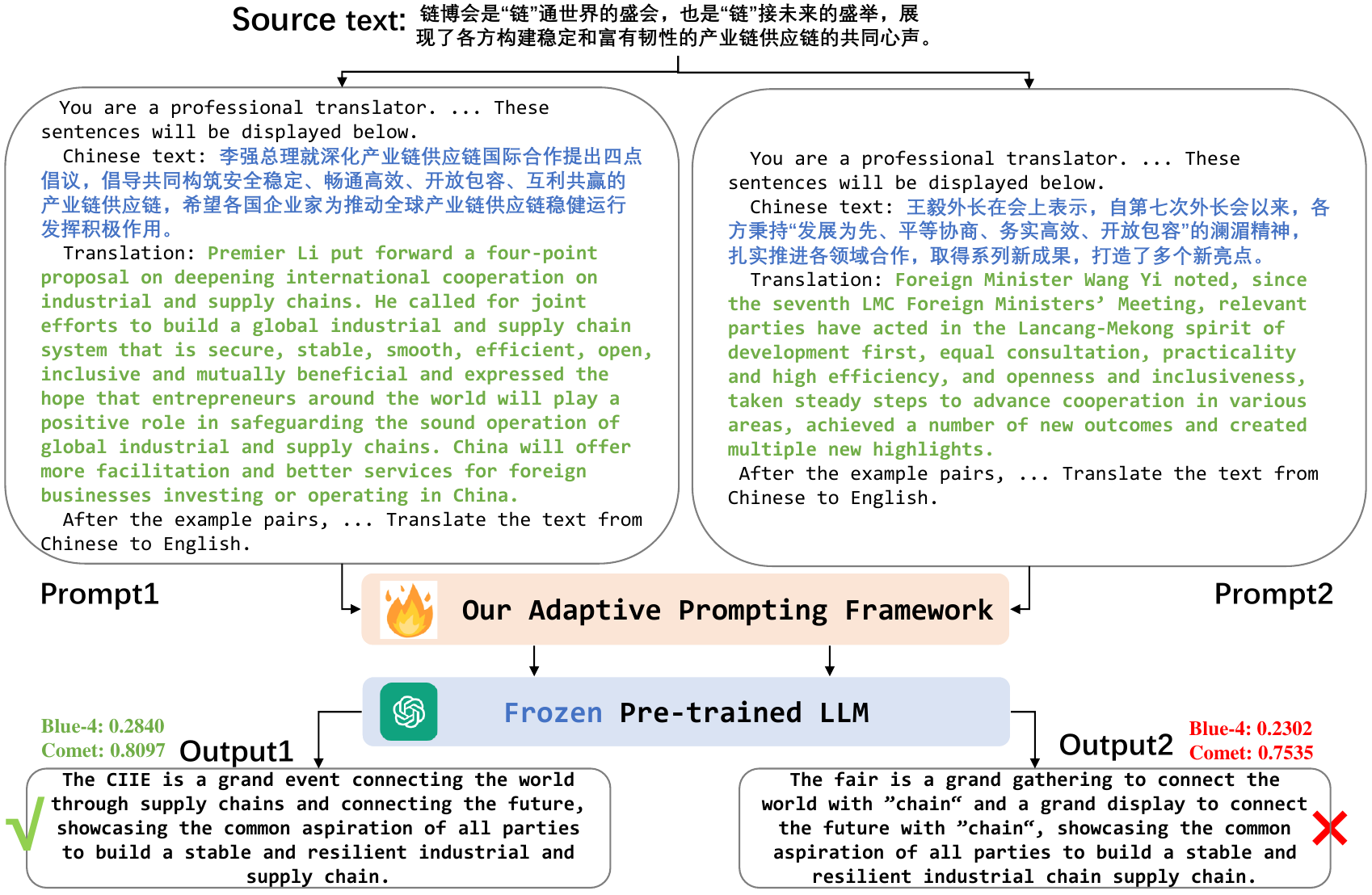}
\caption{Illustration of translation results from different prompts using Llama3-8B. Our adaptive prompting framework can adaptively select the suitable prompt for an input text, yielding a better translation.}
\label{fig_1}
\end{figure}

Pioneering work~\cite{zhang2023prompting} conducted a systematic study on prompting strategies for machine translation with the testbed GLM-130B~\cite{zeng2022glm}, including zero-shot prompting and few-shot prompting. Coincidentally, another work~\cite{zhang-etal-2023-machine} evaluated 15 publicly available language models on machine translation tasks with zero-shot prompting and few-shot learning. \cite{zhu2024multilingual} explored the multilingual translation capabilities of eight popular LLMs, including ChatGPT and GPT-4. Although all these existing works have shown promising translation performance under the settings of both zero-shot prompting and few-shot ICL, they found that the prompt examples matter the translation performance, which means that LLMs are prompt-sensitive. Using suboptimal examples or instructions can degenerate translation. For example, as shown in Figure~\ref{fig_1}, the LLM with prompt 1 which is more related to the input text can generate a better translation result than the LLM with prompt 2 according to the BLEU and Comet. In terms of semantic consistency, we can also observe that the translation quality of the LLM with prompt 1 is higher than the LLM with prompt 2. Therefore, selecting suitable adaptive translation demonstrations to elicit the translation capability of an LLM is crucial for high-quality machine translation under in-context learning. 

Choosing suitable translation demonstrations for different input text is challenging and nontrivial. To address this issue, we propose an \textbf{A}daptive \textbf{F}ew-\textbf{S}hot \textbf{P}rompting (AFSP) framework to automatically select suitable translation demonstrations for various source input sentences to further elicit the translation capability of an LLM for better machine translation. First, we build a translation demonstration retrieval module based on LLM's embedding to retrieve top-k semantic-similar translation demonstrations from aligned parallel translation corpus. The retrieval top-k translation demonstrations will be filled into the handcrafted instruction prompt template which is used for various source sentences uniformly. These translation demonstrations are crucial in eliciting the translation capability of an LLM to generate more semantic-consistent target sentences with current input source sentences. M3-Embedding~\cite{chen2024bge} shows that conducting semantic retrieval with a combination of different retrieval functionalities can achieve better retrieval performance by improving the discrimination of embeddings. Inspired by this, we construct a demonstration retrieval module based on dense embedding, sparse embedding, and multi-vector embedding to build better input representation for retrieving more semantic-related translation demonstrations. The dense embedding, sparse embedding, and multi-vector embedding of a sentence are generated from deployed LLM which is also used for machine translation. Then, we use a constructed adaptive few-show prompt to obtain the translation result in the target language. There is output diversity in an LLM~\cite{kirk2023understanding} due to the probabilistic sampling. Different outputs can lead to different translation quality. To mitigate semantic bias caused by LLMs' probabilistic sampling and ensure semantic-consistent translation, we force the deployed LLM to generate multiple output candidates in the target language and rerank these candidates by a rerank model based on a small language model (SLM). Since there is no available large-scale annotated corpus about the translation quality of different translation outputs and annotating such a corpus is costly, we train the rerank model at a lower cost with a self-supervision way by negative sampling with different text perturbation. With the rerank model, we can choose better translation results, ensuring better semantic consistency between source inputs and target outputs.

Besides, Language evolves throughout time. To better evaluate the effectiveness of our AFSP framework on the latest language and extend the research boundary of neural machine translation, we construct a high-quality diplomatic Chinese-English parallel dataset that consists of 5,528 parallel Chinese-English sentences about the question answers with Chinese foreign-ministry spokesman and foreign journalists. These parallel sentences have very high semantic consistency since they are diplomatically oriented and have been rigorously vetted and proofread. Extensive experiments on our proposed diplomatic Chinese-English parallel dataset and United Nations Parallel Corpus (Chinese-English part) show that the 
effectiveness and superiority of our AFSP.

The main contributions of this work are concluded as follows: First, we propose an adaptive few-shot prompting (AFSP) framework to automatically select suitable translation demonstrations for various source input sentences to further elicit the translation capability of an LLM for better machine translation. Second, in our AFSP, rather than using other embedding models for semantic retrieval, we build a hybrid demonstration retrieval module based on the embedding layer of the deployed LLM itself to build better input representation for retrieving more semantic-related translation demonstrations. Third, to recognize better translation results, we build a rerank model trained in a self-supervision way with negative sampling, ensuring better semantic consistency between source inputs and target outputs and mitigating semantic bias caused by LLMs' probabilistic sampling. Finally, we construct a high-quality diplomatic Chinese-English parallel dataset and extensive experiments on it and the United Nations Parallel Corpus (Chinese-English part) show the effectiveness and superiority of our proposed AFSP quantitatively and qualitatively.

\section{Related work}
The emergence of LLMs has shown outstanding potential in the field of machine translation. Unlike traditional neural machine translation methods~\cite{Bahdanau2014NeuralMT, Sennrich2015NeuralMT, Wang2022UnderstandingAI} which need to be trained with a large-scale machine translation dataset, LLMs were trained on general large-scale corpus and could effectively finish downstream machine translation tasks via prompt engineering or in-context learning without extra model tuning. Current research evaluating and improving the machine translation capabilities of LLMs can be included in two lines. The first line focuses on comprehensive evaluations of LLMs under various translation scenarios, including multilingual translation~\cite{Jiao2023IsCA, Hendy2023HowGA},  document-level translation~\cite{Wang2023DocumentLevelMT, Hendy2023HowGA}, low-resource translation~\cite{Jiao2023IsCA, Bawden2023InvestigatingTT}, etc. Another line focuses on designing novel mechanisms to improve the machine translation capabilities of LLMs, including the design of prompt templates~\cite{Zhang2023PromptingLL, Jiao2023IsCA}, demonstration selection for in-context learning~\cite{Zhang2023PromptingLL, Vilar2022PromptingPF, Garca2023TheUE, yao2023more, Merx2024LowResourceMT, jiang2024can}, self-refinement~\cite{Feng2024TEaRIL, feng2024ladder}, agentic workflow~\cite{Wu2024PerhapsBH, Guo2024SiLLMLL}, etc.

\begin{figure*}[t]
\centering
\includegraphics[width=0.75\linewidth]{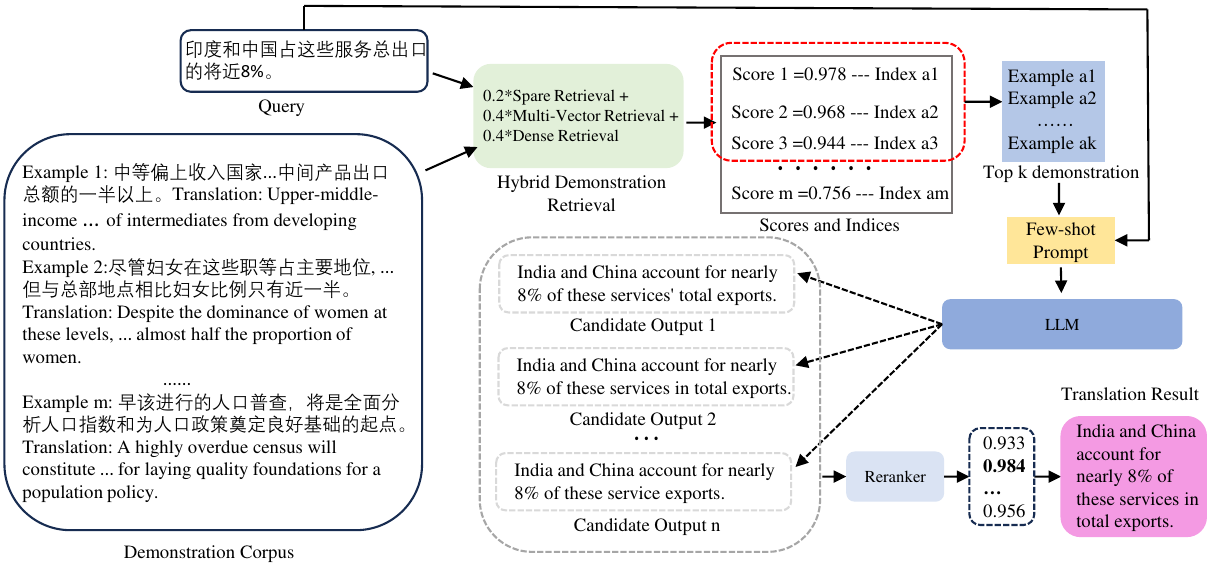}
\caption{The overview of our proposed Adaptive Few-shot Prompting (AFSP) framework.}
\label{fig_method}
\end{figure*}

Among these research lines, the most relevant to our work is the demonstration selection. \cite{Vilar2022PromptingPF} investigated various strategies for choosing translation examples for few-shot prompting. \cite{Garca2023TheUE} outperformed the best-performing system on the WMT'21 English-Chinese news translation task by only using five random examples of English-Chinese parallel data at inference. Both these two works found that example quality is the most important factor, but random sampling will influence their performances.
\cite{yao2023more} proposed a low-resource LLM prompting technique In-Context Sampling (ICS) to produce confident predictions by optimizing the construction of multiple ICL prompt inputs. Leveraging a novel corpus derived from a Mambai language manual and additional sentences translated by a native speaker, \cite{Merx2024LowResourceMT} examine the efficacy of few-shot prompting for machine translation (MT) in the low-resource context by prompting with the strategic selection of parallel sentences and dictionary entries, enhancing translation accuracy.

Different from them, our AFSP automatically selects suitable translation demonstrations for various source input sentences to elicit the translation capability of an LLM for better machine translation. Rather than using other embedding models for semantic demonstration retrieval, our AFSP first deploys a translation demonstration retrieval module based on the deployed LLM's embedding to retrieve top-k semantic-similar translation demonstrations from aligned parallel translation corpus. Then, to ensure better semantic consistency between source inputs and target outputs, we force the deployed LLM to generate multiple output candidates in the target language with the help of translation demonstrations and rerank these candidates with an SLM which is trained in a self-supervised way.

\section{Adaptive Few-shot Prompting (AFSP)}
We introduce our \textbf{A}daptive \textbf{F}ew-\textbf{S}hot \textbf{P}rompting framework, which first adaptively retrieves suitable demonstrations to fill into the placeholder in the prompt template from the demonstration corpus and then sorts the multiple candidate sampled outputs generated by the deployed LLM to obtain final translation result. In this work, except for the demonstration placeholder, we use fixed prompt words in the prompt template for general task description. The overview of the inference phase in our AFSP framework is shown in Figure~\ref{fig_method}. AFSP relies on three key components: a translation demonstration corpus, a hybrid demonstration retrieval module based on the deployed LLM-driven embedding, and a re-ranker module. The translation demonstration corpus aims to provide high-quality parallel translation pairs. The retrieval module takes charge of selecting suitable demonstrations to fill into the prompt template for each input source text. The hybrid demonstration retrieval module is train-free and produces a relevance score based on multiple types of embedding ways for each input source and the text in the demonstration corpus by the deployed LLM for machine translation, rather than using third-party embedding models. With the retrieval demonstrations, we fill the demonstrations into a predefined few-shot prompt and enter it into an LLM for multiple candidate output generation. Finally, we deploy a re-ranker module, which is a small language model (SLM) trained in a self-supervised manner, to sort the generated candidate outputs and obtain the final translation result.

\begin{table}[t]
\centering
\resizebox{1.0\linewidth}{!}{
\begin{tabular}{p{9cm}l}
\hline
You are a professional translator. I will give you one or more examples of text fragments, where the first one is in $ \left\{ src\ lang \right \}$ and the second one is the translation of the first fragment into $ \left\{ tgt\ lang \right \}$. These sentences will be displayed below. \\
1. $ \left\{ src\ lang \right \}$ text: $ \left\{ src\ demo\ 1 \right \}$  \\
$ \left\{ tgt\ lang \right \}$ translation: $ \left\{ tgt\ demo\ 1 \right \}$ \\
2. $ \left\{ src\ lang \right \}$ text: $ \left\{ src\ demo\ 2 \right \}$  \\
$ \left\{ tgt\ lang \right \}$ translation: $ \left\{ tgt\ demo\ 2 \right \}$ \\
3. $ \left\{ src\ lang \right \}$ text: $ \left\{ src\ demo\ 3 \right \}$  \\
$ \left\{ tgt\ lang \right \}$ translation: $ \left\{ tgt\ demo\ 3 \right \}$ \\
...\\
k. $ \left\{ src\ lang \right \}$ text: $ \left\{ src\ demo\ k \right \}$  \\
$ \left\{ tgt\ lang \right \}$ translation: $ \left\{ tgt\ demo\ k \right \}$ \\
After the example pairs, I will provide a/an $ \left\{ src\ lang \right \}$ sentence and I would like you to translate it into $ \left\{ tgt\ lang \right \}$. Please provide only the translation result without any additional comments, formatting, or chat content. Translate the text from $ \left\{ src\ lang \right \}$ to $ \left\{ tgt\ lang \right \}$. \\
\hline
\end{tabular}
}
\caption{Illustration of the few-shot prompt used in our work. The placeholders $ \left\{ src\ lang \right \}$ and $ \left\{ tgt\ lang \right \}$ will be replaced with specific source and target language names in practice, such as Chinese, English, etc. Similarly, the placeholders $ \left\{ src\ demo\, i \right \}$ and $ \left\{ tgt\ demo\, i \right \}$ will also be replaced with retrieved parallel translation demonstrations where $i \in [1, ..., k]$ .}
\label{prompt_template}
\end{table}

\subsection{Prompt template and Demonstration Corpus}
In AFSP, as shown in Table~\ref{prompt_template}, we only use a fixed prompt template with variable placeholders inspired from prior works~\cite{Jiang2024ConvergencesAD, Agarwal2024ManyShotIL}. We do not focus on the diversified design of prompt templates in this work and achieve adaptive prompts for different source text by filling suitable demonstrations according to the retrieved results from the demonstration corpus. Demonstration corpus can be any high-quality parallel translation corpus. In practice, it can be expanded with extra parallel translation corpus. For the sake of simplicity, we simply use the training set from specific translation tasks as the demonstration sources to build the demonstration corpus. For example, we use the training part in the UN Open Corpus v1.0 (Chinese and English versions) as the demonstration corpus when conducting Chinese-English bilingual translation. Similarly, we also use the training subset of our newly constructed Diplomatic corpus as the demonstration corpus when conducting machine translation on its test set. 

\subsection{Hybrid Demonstration Retrieval}
As claimed in the pioneering works~\cite{zhang2023prompting, zhang-etal-2023-machine, zhu2024multilingual}, the number, quality, and semantic similarity of prompt examples matter the translation performance. Therefore, it is crucial to adaptively retrieve high-quality and highly semantic similar demonstrations for different input texts to achieve better LLM prompting for better eliciting the translation capability of an LLM. To achieve this goal, superior embedding-based semantic representation is essential. \cite{chen2024bge} shows that a combination of different embedding-based retrieval functionalities can improve the discrimination of embedding-based semantic representation. Inspired by them, we build an embedding-based hybrid demonstration retrieval module for demonstration retrieval in a training-free way by utilizing the embedding matrix of the deployed LLM that conducts machine translation. The retrieval results are sorted by a weighted combination of relevance scores based on dense embedding, sparse embedding, and multi-vector embedding. The reason we use the deployed LLM as the embedding generator rather than other embedding models is that LLM is pre-trained with large-scale general corpus and can represent text accurately.

Formally, given a query text $q$ in the source language, the demonstration retrieval module can retrieve translation demonstration $\left \langle d^{src}, d^{tgt} \right \rangle$ from the corpus $\mathcal{D}$ based on the hybrid relevance score $s_rank$ of $q$ and $d^{src}$: $\left \langle d^{src}, d^{tgt} \right \rangle = f_h(q, \mathcal{D})$. Here, $f_h(\cdot)$ denotes the retrieval function based on the hybrid relevance score. For the text $q$, dense embedding, sparse embedding, and multi-vector embedding can be formalized separately as follows: 1) dense embedding $e^{dense}_q$: the text $q$ is first transformed into the embedding vectors $\textbf{E}_q$ based on the embedding layer in the LLM. Then, we obtain $e^{dense}_q$ by conducting max pooling on $\textbf{E}_q$ and normalization: $e^{dense}_q = norm(MaxPooling(\textbf{E}_q))$. 2) sparse embedding $e^{sparse}_q$: the embedding vector $\textbf{E}_q$ is also used to estimate the importance of each token to facilitate lexical representation. For each token $t$ within the text $q$, the token weight is calculated as $w_{q_t} = ReLU(\textbf{W}_{sparse}^T\textbf{E}_q[t])$, where $ReLU$ is rectified linear unit and $ \textbf{W}_{sparse} \in \mathbb{R}^{H\times1}$. $H$ is the dimension size of the embedding and $\textbf{W}_{sparse}$ is a projection matrix mapping token embedding into a float number as its importance. It is only initialized by Gaussian initialization since we found Gaussian initialization is enough to make the model work fine without any model training. 3) multi-vector embedding $e^{multi}_q$: it is an extension of dense embedding by utilizing the entire output embeddings for text representation: $e^{multi}_q = norm(\textbf{W}_{multi}^T\textbf{E}_q)$, where $\textbf{W}_{multi} \in \mathbb{R}^{H\times H}$ is a projection matrix initialized by Gaussian initialization. 

With the above three embeddings with different granularities, we can calculate three relevance scores for multi-granularity retrieval. For dense retrieval, given a text $q$ and source demonstration $p$, we can compute the relevance score $s_{dense}$ by the inner product between the two embeddings $e^{dense}_q$ and $e^{dense}_p$ as follows: $s_{dense} = e^{dense}_q \cdot e^{dense}_p$. For sparse retrieval, we can compute $s_{sparse}$ by the joint importance of the co-existed tokens (denoted as $q \cap p$) as follows: $s_{sparse} = \sum_{t\in q \cap p}(w_{q_t} \times w_{q_t})$. For multi-vector retrieval, we can compute $s_{multi}$ by late interaction as follows: $s_{sparse} = \frac{1}{l_q}\sum_{i=1}^{l_q} max_{j=1}^{l_p} e^{multi}_q[i]\cdot {e^{multi}_p[j]}^T$, where $l_q$ and $l_p$ are the lengths of text $q$ and source demonstration $p$.

Based on the above three relevance scores, we conduct the demonstration retrieval in a hybrid process according to $s_{rank}$ which can be defined as follows:
\begin{equation}
s_{rank} = \alpha_1 \times s_{dense} + \alpha_2 \times s_{sparse} + \alpha_3 \times s_{multi}
\end{equation}
where $\alpha_1$, $\alpha_2$, and $\alpha_3$ are three hyper-parameters to adjust the weights of three retrieval functionality.

\subsection{Result Re-ranking}
There is output diversity in an LLM~\cite{kirk2023understanding} due to the probabilistic sampling. The different outputs may have different semantic biases, which will influence the final translation performance. To mitigate this issue, we force the deployed LLM to generate multiple output candidates in the target language and rerank these candidates by a re-ranker model based on a small language model (SLM). The re-ranker takes charge of scoring the output candidates. However, training this re-ranker is challenging since there is available large-scale annotated corpus about the translation quality of different translation outputs and annotating such a corpus is costly. Therefore, we design a self-supervised training method to train such a re-ranker at a low cost by conducting negative sampling with different text perturbations.  

\subsubsection{Negative Sampling} Formally, given the parallel translation corpus $\mathcal{D} = \left\{ <d^{src}_1, d^{tgt}_1>, <d^{src}_2, d^{tgt}_2>, ..., <d^{src}_N, d^{tgt}_N> \right\}$ where $d^{src}_i$ and $d^{tgt}_i$ are the texts in the source language and the target language respectively. To construct a dataset $\mathcal{D}' = \left\{ <d^{tgt'}_1, s'_1>, <d^{tgt'}_2, s'_2>, ..., <d^{tgt'}_M, s'_M> \right\}$ to train the re-ranker, we can disturb the $d^{tgt}_i$ by multiple degeneration operation set $A$ including converting to the parallel text ($Parallel$), back translation ($Back$), inserting source text ($Insert$), spelling mistake ($Se$), repeated translation ($Ret$), synonym replacement ($Replace$). $d^{tgt'}_i$ and $s'_i$ are the degenerated text and corresponding quality score, respectively. We define the quality score of the original target text $d^{tgt}_i$ as 1. Assuming that $B$ contains a null operation that means we just copy the original text into $\mathcal{D}'$ and all possible combinations of the degeneration operation in $A$, for each possible combination $b_i \in B$, we can obtain the degenerated text $d^{tgt'}_i$ and calculate its score $s'$ as follows:
\begin{equation}
\begin{aligned}
\label{s}
    d^{tgt'}_i &= f_{b_{i}}(d^{tgt}_i), b_i \in B \\
    s'_i &= 1 - 0.2 \cdot| b_i |
\end{aligned}
\end{equation}
where $f_{b_i}(\cdot)$ represents the degeneration function with the degenration operation combination $b_i$ and $|b_i|$ is the number of degeration operations in $b_i$.  In this way, we can generate a large-scale dataset $\mathcal{D}'$ from the parallel translation corpus $\mathcal{D}$ to train the re-ranker in a self-supervised manner.

\subsubsection{Re-ranker and Learning Objectives}
We deploy BERT~\cite{devlin-etal-2019-bert} as the backbone of the SLM in the Re-ranker. For Chinese-English translation, we use Bert-large-cased\footnote{https://huggingface.co/google-bert/bert-large-cased} while we use  Bert-based-Chinese\footnote{https://huggingface.co/google-bert/bert-base-chinese} as the SLM for English-Chinese translation. Given a degenerated text $d^{tgt'}_i$ and its quality score $s'_i$, the re-ranker takes the degenerated text $d^{tgt'}_i$ as input and predicts a quality assessment score $s^{rerank}_i$ as close to the annotated score $s'_i$ as possible. The re-ranker calculates the quality assessment score by applying a Linear layer to map the output encoding of [CLS] token into 1-D float number followed by the $Sigmoid$ function to normalize the output to between 0 and 1. To optimize the re-ranker, we adopted Mean Squared Error as its objective function to enable the re-ranker to predict quality assessment scores. Therefore, the re-ranker and its learning objective can be modeled as follows:
\begin{equation}
\begin{aligned}
\textbf{E} &= BERT(d^{tgt'}_i) \\
s^{rerank}_i &= Sigmoid(Linear(\textbf{E}[0])) \\
\mathcal{L} &= \left\| s^{rerank}_i - s'_i \right\|_2
\end{aligned}
\end{equation}

\section{Experiments}

\subsection{Experiment Settings}
\subsubsection{Datasets}
To validate the effectiveness of the proposed AFSP, we first crawled a high-quality parallel Chinese-English dataset named Diplomatic corpus from the China Diplomatic website\footnote{https://www.fmprc.gov.cn/}. The Diplomatic corpus consists of speeches made by spokespersons during routine press conferences, including questions posed by journalists and responses from Chinese spokespersons on a range of diplomatic issues. There are some advantages in the Diplomatic corpus. The first is accessibility. All data is publicly available on the China Diplomatic website and can be easily found online. The second is high quality. All bilingual materials are translated by professional translators from specialized institutions, ensuring superior quality. The third is language complexity. The Diplomatic corpus contains certain political terminology and specialized terms, which may pose challenges for the LLM in the context of China's political landscape. The final one is recency. All texts are recorded from 2022 to 2023 which can reflect the latest advances in language.
Besides, we also use a Chinese-English subset from the UN Open Corpus v1.0 as the second testbed, which can show the universality of the proposed AFSP. We use UN to denote this subset. The UN Parallel Corpus is a parallel corpus that includes official UN documents and statements from meetings. The content covers various fields such as politics, economics, culture, and technology. This corpus records texts written and manually translated from 1990 to 2014, aligned at the sentence level. We randomly selected 120,000 entries as the data for evaluation.
The statistics for both two datasets are shown in Table~\ref{tabel22}. For both two datasets Diplomatic and UN, we randomly selected 500 parallel translation pairs to serve as the test set for evaluating AFSP. The remaining pairs are used as the demonstration corpus for adaptive demonstration retrieval. 

\begin{table}[t]
\centering
\resizebox{0.8\linewidth}{!}{
\begin{tabular}{c|c|c|c|cc}
\hline
 Dataset&  Language& \#Sent. &\#Word  & \#Average. \\
 \hline
\multirow{2}{*}{Diplomatic} &  English&  5.528K& 415K & 75.20 \\
 \cline{2-5}
 & Chinese & 5.528K & 316K & 57.29 \\
 \hline
 \multirow{2}{*}{UN}& English & 120K & 3,500K &29.11  \\
 \cline{2-5}
 &  Chinese& 120K & 3,220K &26.75 \\
 \hline

\end{tabular}}

 \caption{Statistics of the Diplomatic Corpus and UN. The statistics include the number of sentences (\#Sent.), the number of words (\#Word), and the average number of words per sentence (\#Average.). Both datasets consist of sentences of varying lengths.}
 \label{tabel22}
\end{table}

\begin{table*}[!t]
\centering
\resizebox{\linewidth}{!}{
\begin{tabular}{c|ccccccc|ccccccc}
\hline
\multicolumn{15}{c}{\textbf{ChatGLM3-6B}}  \\    
\hline
\multirow{2}{*}{Methods}  & \multicolumn{7}{c|}{English-to-Chinese} & \multicolumn{7}{c}{Chinese-to-English}\\
\cline{2-15}
& BLUE-4 & METEOR &ROUGE-1 &ROUGE-2 &ROUGE-L & CHRF & COMET-Kiwi & BLUE-4 &METEOR &ROUGE-1 &ROUGE-2  &ROUGE-L & CHRF  & COMET-Kiwi  \\
\hline
Zero-shot& 18.89 &50.57 & 21.23 &  10.25& 21.13 & 42.76 & 88.07 & 22.69 & 54.64 & 59.91 & 31.99 & 50.93 & 66.50 & 81.47 
\\
\hline
Few-shot&20.27 &51.73 &20.54 &10.28 &20.46 &43.81 &88.22 & 23.61&55.70  &60.84  &33.05 &51.63  &67.06 &81.39\\
\hline
KNN Few-shot&20.22 &51.77 &20.77 &10.36 &20.68 &43.90 &88.33& 22.93&55.14  &60.40  &32.50 &51.13&66.69 &81.25  \\
\hline
AFSP w/o rerank (Ours) &25.92 &56.80 &22.47 &11.39 &22.40 &48.58 &88.99 &28.14 &59.29 &63.83 &37.32 &54.94 &69.34 &82.32 \\
\hline
AFSP (Ours)& \textbf{27.36} &\textbf{58.22}  &\textbf{22.92}  &\textbf{11.52  }&\textbf{22.85}  &\textbf{49.78}  &\textbf{89.12} &\textbf{29.17} & \textbf{60.01} &\textbf{64.02}  & \textbf{37.64} &\textbf{ 55.15} & \textbf{69.68} & \textbf{82.74}\\
\hline

\multicolumn{15}{c}{\textbf{InternLM2-7B}}  \\    
\hline
 \multirow{2}{*}{Methods} & \multicolumn{7}{c|}{English-to-Chinese} & \multicolumn{7}{c}{Chinese-to-English}\\
\cline{2-15}
& BLUE-4 & METEOR &ROUGE-1 &ROUGE-2 & ROUGE-L& CHRF  & COMET-Kiwi & BLUE-4 & METEOR &ROUGE-1 &ROUGE-2 & ROUGE-L & CHRF  & COMET-Kiwi  \\
\hline
Zero-shot& 20.60 &52.32 &22.41 &11.14 &22.25 &44.49 &88.42&23.36 &56.06 &60.43 &32.62 &51.16 &67.72 &81.92\\
\hline
Few-shot&24.45 &55.69 &22.66 &11.64 &22.51 &47.21 &89.01&24.15 &57.06 &61.40 &33.81 &52.24 &68.48 &81.66  \\
\hline
KNN Few-shot&24.91 &56.28 &22.81 &11.71 &22.64 &47.61 &88.95&24.73 &57.42 &61.68 &34.33 &52.63 &68.66 &81.94 \\
\hline
AFSP w/o rerank (Ours) &30.72 &60.78 &22.86 &11.92 &22.72 &52.28 &89.52 &30.10 &61.24 &64.98 &39.34 &56.28 &71.01 &82.53 \\
\hline
AFSP (Ours)&\textbf{31.75} &\textbf{61.34} &\textbf{22.86} &\textbf{11.93} &\textbf{22.71} &\textbf{52.85} &\textbf{89.62}&\textbf{31.28} &\textbf{61.90} &\textbf{64.87} &\textbf{39.66} &\textbf{56.09} &\textbf{70.87} &\textbf{82.74}  \\
\hline

\multicolumn{15}{c}{\textbf{Llama3-8B}}  \\    
\hline
 \multirow{2}{*}{Methods}  & \multicolumn{7}{c|}{English-to-Chinese} & \multicolumn{7}{c}{Chinese-to-English} \\
\cline{2-15}
& BLUE-4 & METEOR & ROUGE-1 &ROUGE-2 &ROUGE-L & CHRF & COMET-Kiwi & BLUE-4 & METEOR & ROUGE-1 &ROUGE-2 &ROUGE-L & CHRF  & COMET-Kiwi\\
\hline
Zero-shot&10.72 &37.16 &17.44 &8.12 &17.22 &32.02 &84.57&24.03 &56.68 &60.91 &33.55 &51.66 &68.21 &82.27  \\
\hline
Few-shot&15.53 &45.71 &18.51 &8.83 &18.33 &38.97 &86.22&25.66 &57.98 &62.40 &35.18 &53.13 &68.83 &82.81\\
\hline
KNN Few-shot&16.51 &46.81 &18.69 &8.98 &18.54 &39.90 &86.52&25.32 &57.58 &62.27 &35.07 &52.99 &68.41 &82.76 \\
\hline
AFSP w/o rerank (Ours) &26.34 &55.55 &19.81 &9.56 &19.60 &47.95 &87.96 &30.71 &61.46 &65.72 &40.24 &57.10 &71.10 &83.29 \\
\hline
AFSP (Ours)&\textbf{27.67} &\textbf{56.71} &\textbf{19.86} &\textbf{9.87} &\textbf{19.65} &\textbf{49.05} &\textbf{88.41}&\textbf{31.02} &\textbf{61.48} &\textbf{65.73} &\textbf{40.25} &\textbf{56.97} &\textbf{70.98} &\textbf{83.34} \\
\hline

\multicolumn{15}{c}{\textbf{Chatgpt-3.5-turbo-0125}}  \\    
\hline
 \multirow{2}{*}{Methods} & \multicolumn{7}{c|}{English-to-Chinese} & \multicolumn{7}{c}{Chinese-to-English}  \\
\cline{2-15}
& BLUE-4 & METEOR & ROUGE-1 &ROUGE-2 &ROUGE-L & CHRF & COMET-Kiwi & BLUE-4 & METEOR & ROUGE-1 &ROUGE-2 &ROUGE-L & CHRF  & COMET-Kiwi \\
\hline
Zero-shot&22.96 &54.90 &22.34 &11.13 &22.22 &46.54 &89.13&27.66 &59.92 &64.11 &37.29 &55.47 &70.65 &83.24\\
\hline
Few-shot&24.57 &56.73 &22.33 &11.17 &22.18 &48.18 &89.51&28.12 &60.54 &64.55 &37.71 &55.98 &70.96 &83.43 \\
\hline
KNN Few-shot&25.63 &57.83 &22.42 &11.38 &22.33 &49.09 &89.49&28.34 &60.49 &64.53 &37.91 &56.00 &70.89 &83.45 \\
\hline
AFSP w/o rerank (Ours) &30.26 &62.10 &23.47 &11.73 &23.36 &52.93 &89.94 &31.29 &62.46 &66.42 &40.75 &58.00 &72.12 &83.86 \\
\hline
AFSP (Ours)&\textbf{32.30} &\textbf{63.53} &\textbf{23.26} &\textbf{11.69} &\textbf{23.26} &\textbf{54.31} &\textbf{90.32} &\textbf{32.30} &\textbf{63.19} &\textbf{66.93} &\textbf{41.44}0&\textbf{58.67} &\textbf{72.46} &\textbf{83.95}\\
\hline

\end{tabular}
}

\caption{Performance Comparison on Diplomatic Corpus. The best result is highlighted in \textbf{bold}.} 
\label{main_ex}
\end{table*}

\begin{table*}[!t]
\centering
\resizebox{\linewidth}{!}{
\begin{tabular}{c|ccccccc|ccccccc}
\hline
\multicolumn{15}{c}{\textbf{ChatGLM3-6B}}  \\    
\hline
\multirow{2}{*}{Methods} & \multicolumn{7}{c|}{English-to-Chinese} & \multicolumn{7}{c}{Chinese-to-English}\\
\cline{2-15}
& BLUE-4 & METEOR & ROUGE-1 &ROUGE-2 &ROUGE-L & CHRF & COMET-Kiwi & BLUE-4 & METEOR & ROUGE-1 &ROUGE-2 &ROUGE-L & CHRF  & COMET-Kiwi\\
\hline
Zero Few-shot& 18.89 & 52.89 &42.15  &  17.93& 41.72 &45.21 &86.94 &21.46&58.02&59.99&34.20&53.20&64.11&83.64
\\
\hline
Few-shot& 19.05&52.62  &42.14  &17.62  &41.64 &45.07  &86.33 & 21.75 & 58.21 & 59.43&34.57 &52.82&65.25&82.46\\
\hline
KNN Few-shot& 18.80 &52.29  &41.63  &17.50  &41.24  &44.89  &86.56  &22.74  &59.01  &60.07&35.33&53.46&65.12&83.08  \\
\hline
AFSP w/o rerank (Ours) &\textbf{24.84} &57.28 &\textbf{42.83}	&18.34 &\textbf{42.45} &50.15 &87.99 &\textbf{29.35} &63.44 &64.59 &41.67 &58.22 &69.00 &84.55 \\
\hline
AFSP (Ours)& 24.61 &\textbf{57.60}  &42.80  &\textbf{18.34}&42.44&\textbf{50.52}&\textbf{88.24}&29.07 &\textbf{64.49} &\textbf{65.08}  &\textbf{41.75}  &\textbf{58.39}  &\textbf{69.64} &\textbf{84.87}  \\
\hline

\multicolumn{15}{c}{\textbf{InternLM2-7B}}  \\    
\hline
\multirow{2}{*}{Methods}  & \multicolumn{7}{c|}{English-to-Chinese} & \multicolumn{7}{c}{Chinese-to-English} \\
\cline{2-15}
& BLUE-4 & METEOR & ROUGE-1 &ROUGE-2 &ROUGE-L & CHRF & COMET-Kiwi & BLUE-4 & METEOR & ROUGE-1 &ROUGE-2 &ROUGE-L & CHRF  & COMET-Kiwi\\
\hline
Zero Few-shot& 17.32 &50.31 &37.36 &17.82 &36.81 &43.49 &86.30&25.48 &61.86 &63.52 &38.68 &56.96 &67.23 &84.71\\
\hline
Few-shot&17.85 &51.53 &41.13 &18.39 &40.70 &44.43 &86.70&25.76 &62.65 &63.49 &38.81 &57.00 &68.35 &83.92  \\
\hline
KNN Few-shot&17.80 &51.27 &39.45 &18.16 &39.04 &44.30 &86.58&26.30 &62.61 &63.95 &39.27 &57.28 &68.02 &84.45 \\
\hline
AFSP w/o rerank (Ours) &\textbf{26.21} &58.49 &40.45 &19.17 &40.13 &51.59 &88.34 &31.28 &61.90 &64.87 &39.66 &56.09 &\textbf{72.84} &86.12 \\
\hline
AFSP (Ours)& 26.10 & \textbf{58.76} & \textbf{40.46} & \textbf{19.27} & \textbf{40.13} & \textbf{51.75} & \textbf{88.91} & \textbf{33.98} & \textbf{67.37} & \textbf{68.63} & \textbf{46.27} & \textbf{62.47} & 71.93 & \textbf{86.50} \\
\hline

\multicolumn{15}{c}{\textbf{Llama3-8B}}  \\    
\hline
\multirow{2}{*}{Methods}   & \multicolumn{7}{c|}{English-to-Chinese} & \multicolumn{7}{c}{Chinese-to-English}  \\
\cline{2-15}
& BLUE-4 & METEOR & ROUGE-1 &ROUGE-2 &ROUGE-L & CHRF & COMET-Kiwi & BLUE-4 & METEOR & ROUGE-1 &ROUGE-2 &ROUGE-L & CHRF  & COMET-Kiwi\\
\hline
Zero Few-shot&8.85 &35.40 &34.21 &15.43 &33.66 &30.26 &81.12&18.60 &55.35 &58.18 &32.07 &51.08 &63.92 &83.38 \\
\hline
Few-shot&13.11 &44.47 &34.05 &15.30 &33.39 &38.42 &83.41&21.91 &58.12 &60.77 &35.13 &53.67 &65.88 &84.25\\
\hline
KNN Few-shot&12.64 &43.78 &29.44 &15.04 &28.89 &37.87 &83.47&21.72 &57.58 &60.66 &35.14 &53.60 &65.73 &84.28 \\
\hline
AFSP w/o rerank (Ours) &22.74 &53.99 &33.11	&16.46 &32.54 &47.79 &86.51 &30.11 &63.94 &66.75 &44.22 &60.69 &71.31 &\textbf{86.26} \\
\hline
AFSP (Ours)& \textbf{23.62} & \textbf{54.53} & \textbf{34.26} & \textbf{16.90} & \textbf{33.84} & \textbf{48.45} & \textbf{87.21} & \textbf{31.03} & \textbf{64.88} & \textbf{67.06} & \textbf{44.94} & \textbf{60.95} & \textbf{71.68} & 86.24\\
\hline

\multicolumn{15}{c}{\textbf{Chatgpt-3.5-turbo-0125}}  \\    
\hline
\multirow{2}{*}{Methods}   & \multicolumn{7}{c|}{English-to-Chinese} & \multicolumn{7}{c}{Chinese-to-English}\\
\cline{2-15}
& BLUE-4 & METEOR & ROUGE-1 &ROUGE-2 &ROUGE-L & CHRF & COMET-Kiwi & BLUE-4 & METEOR & ROUGE-1 &ROUGE-2 &ROUGE-L & CHRF  & COMET-Kiwi\\
\hline
Zero Few-shot&20.04 &52.94 &31.96 &17.48 &31.51 &46.28 &87.10&24.49 &61.31 &63.27 &37.99 &56.90 &67.80 &85.43\\
\hline
Few-shot&21.29 &55.39 &42.96 &18.62 &42.51 &47.93 &87.88&27.04 &63.92 &65.33 &40.59 &58.91 &69.65 &85.90\\
\hline
KNN Few-shot&20.27 &53.80 &35.09 &17.95 &34.66 &47.05 &87.51 &27.10 &63.13 &64.90 &40.59 &58.65 &69.30 &85.85 \\
\hline
AFSP w/o rerank (Ours) &28.41 &61.09 &40.71 &18.98 &40.33 &53.92 &89.10 &32.74 &66.86 &68.85 &46.24 &63.14 &72.99 &86.91 \\
\hline
AFSP (Ours)& \textbf{29.06} & \textbf{62.00} & \textbf{42.40} & \textbf{19.05} & \textbf{42.00} & \textbf{54.59} & \textbf{89.48} & \textbf{34.09} & \textbf{67.48} & \textbf{69.28} & \textbf{46.96} & \textbf{63.65} & \textbf{73.32} & \textbf{87.34} \\
\hline
\end{tabular}
}
\caption{Performance Comparison on UN. The best result is highlighted in \textbf{bold}.}
\label{main_ex1}
\end{table*}

\subsubsection{Metrics}
To conduct a comprehensive assessment of translation quality, we employed the most commonly used BLEU~\cite{Papineni2002BleuAM}, METEOR~\cite{Banerjee2005METEORAA}, ROUGE-1, ROUGE-2, ROUGE-L~\cite{Lin2004ROUGEAP}, CHRF~\cite{popovic-2015-chrf}, and COMET-Kiwi\cite{comet2022} as evaluation metrics. Besides, we also evaluate our AFSP by conducting a human evaluation of fluency, accuracy, and consistency of style. 

\subsubsection{Implementation Details}
We evaluate AFSP on three open-source LLMs, ChatGLM3-6B, InternLM2-7B, and Llama3-8B, as well as one closed-source LLM  ChatGPT-3.5-turbo-0125 by comparing with three baselines, Zero-shot~\cite{jiang2024can}, Few-shot~\cite{jiang2024can}, kNN-based few-shot~\cite{nori2023generalistfoundationmodelsoutcompete} in both Chinese-to-English and English-to-Chinese translation directions. The $\alpha_1$, $\alpha_2$, and $\alpha_3$ are set to 0.4, 0.4, 0.2 for the computation of the final relevance score $s_{rank}$ in hybrid demonstration retrieval. The shot $k$ for few-shot prompts is set to 3 due to the limited GPU memory of NVIDIA RTX 3090. For the closed-source ChatGPT-3.5-turbo-0125, we deploy ChatGLM3-6B as the embedding model for hybrid demonstration retrieval. We conduct top-30 sampling for ChatGLM3-6B, InternLM2-7B, and Llama3-8B and top-5 sampling for ChatGPT-3.5-turbo-0125 due to cost considerations.  

\subsection{Experiment Result}
\subsubsection{Main Results}
Table~\ref{main_ex} and Table~\ref{main_ex1} show the performance of our AFSP and baselines on different translation datasets and different LLMs. Compared to baselines, our AFSP demonstrates superior performance by always generating higher-quality translation according to various metrics.  For instance, when Llama-3-8B translates from Chinese to English on the UN, our ASFP achieves significant improvements over KNN Few-shot with 9.31 improvement in BLEU-4, 7.3 improvement in METEOR, 7.35 improvement in ROUGE-L, and  1.96 improvement in COMET-Kiwi. Other models also show significant metric improvements in translation across different datasets and LLMs, highlighting the effectiveness of our AFSP method.

\begin{table}[t]
\centering
\resizebox{\linewidth}{!}{
\begin{tabular}{c|ccc|ccc}
\hline
Method        & Fluency & Accuracy & Style  & Fluency & Accuracy & Style  \\
\hline
\multicolumn{1}{c|}{}&\multicolumn{3}{c|}{Chinese-to-English} &\multicolumn{3}{c}{English-to-Chinese}\\
\hline
Zero-shot & 0.1428  & 0.2071   & 0.2    & 0.1428  & 0.1214   & 0.1428 \\
Few-shot      & 0.25714 & 0.2571   & 0.2714 & 0.1428  & 0.25     & 0.1714 \\
KNN Few-shot  & 0.2714  & 0.2285   & 0.1928 & 0.1785  & 0.1428   & 0.15   \\
ASFP (Ours)         & \textbf{0.32857 }& \textbf{0.3071}   & \textbf{0.3071} & \textbf{0.5357}  & \textbf{0.4857}   & \textbf{0.5357}\\
\hline
\end{tabular}}

\caption{Human evaluation results on the Chinese-to-English and English-to-Chinese translation performance for the Diplomatic corpus. We calculated the proportion of selections made by the testers. The best results in each aspect are highlighted in \textbf{bold}.}
\label{table:999}

\end{table}

\subsubsection{Human Evaluation}

To further validate the effectiveness of the AFSP, we also conducted a human evaluation of both two datasets and two translation directions to compare AFSP with baselines. For each translation direction, we randomly selected 5 examples from each dataset. Participants judged the options based on fluency, accuracy, and style retention by selecting the sentence they deemed best. We tested the translation results generated by Llama3-8B and invited 14 teachers or students fluent in English or Chinese to participate in the evaluation for each translation direction. To avoid bias, the output order was randomized. The evaluation results in Table~\ref{table:999} demonstrate that our AFSP  outperforms all baselines in fluency, semantic accuracy, and style consistency.

\begin{table}[t]
\centering
\resizebox{\linewidth}{!}{
\begin{tabular}{c|ccccc}
\hline
 Embedding& BGE-M3 & E5-Large & BGE-large &  BCE&  ChatGLM3-6B \\
 \hline
\multicolumn{1}{c|}{}&\multicolumn{5}{c}{Chinese-to-English} \\    
 \hline
 BLUE-4& 27.31 & 26.39 & 22.91 & 27.99 &\textbf{29.17}\\
 METEOR& 58.69 & 58.00 & 54.88 & 59.15 & \textbf{60.02} \\
 ROUGE-1&  63.42& 62.83 & 60.19 & 63.79 &  \textbf{64.02} \\
 ROUGE-2& 36.57 & 35.73 & 32.34 & 37.14 & \textbf{37.65}   \\
 ROUGE-L& 54.45 & 53.80 & 50.87 &  54.90& \textbf{55.15}\\
 CHRF& 69.00 & 68.54 & 66.47 &  69.30& \textbf{69.68}\\
 COMET-Kiwi& 82.18 & \textbf{82.90} & 81.16 & 82.29 & 82.74\\
 \hline
\multicolumn{1}{c|}{}& \multicolumn{5}{c}{English-to-Chinese} \\
\hline
BLUE-4&23.61 & 24.12&20.29 & 24.74& \textbf{27.37}  \\
METEOR&54.85 &55.06&51.80 &55.59& \textbf{58.23 } \\
ROUGE-1&21.46 &21.95&20.61&21.80&\textbf{22.92 } \\
ROUGE-2&10.76&10.90&10.40 &10.81 &\textbf{11.53} \\
ROUGE-L&21.39 &21.86&20.53&21.69&\textbf{22.85} \\
CHRF&46.65&46.97&43.92&47.53&\textbf{49.78 } \\
COMET-Kiwi& 88.73&88.69&88.29&88.85&\textbf{89.12} \\
 \hline
\end{tabular}
}

\caption{The translation performance of ASFP with different embedding models for ChatGLM3-6B on the Diplomatic Corpus dataset. The best result is highlighted in \textbf{bold}.}
\label{tabel:6}
\end{table}

\subsubsection{The Choice of Embedding Model} In our hybrid demonstration retrieval, we use the embedding model in the deployed LLM to compute relevance scores. To show the effectiveness of using the embedding model of the deployed LLM model, we conduct an ablation study on various embedding models including BGE-M3, E5-Large, BGE-large, and BCE. The results in Table~\ref{tabel:6} show using the embedding model of the deployed LLM model is a better choice than using third-party embedding models. 

\begin{table}[t]
\centering
\Huge
\resizebox{\linewidth}{!}{
\begin{tabular}{c|ccccccc}
\hline
 $\alpha_1$/$\alpha_2$/$\alpha_3$& 0.2/0.4/0.4 & 0.3/0.3/0.4 &0.4/0.3/0.3 &  0.25/0.25/0.5&  0.25/0.35/0.4  & 0.35/0.35/0.3 & 0.4/0.4/0.2 \\
 \hline

\multicolumn{1}{c|}{}&\multicolumn{7}{c}{Chinese-to-English} \\    
 \hline
 BLUE-4& 27.77 &27.85 &27.90 &27.87 &27.75 &27.86 &\textbf{28.14}\\
 METEOR& 58.90 &58.99 &59.02 &58.97 &58.95 &58.99 &\textbf{59.29}\\
 ROUGE-1&  63.54 &63.67 &63.70 &63.65 &63.63 &63.67 &\textbf{63.83}\\
 ROUGE-2& 36.97 &37.03 &37.10 &37.10 &37.05 &37.08 &\textbf{37.32} \\
 ROUGE-L& 54.65 &54.70 &54.79 &54.75 &54.71 &54.79 &\textbf{54.94}\\
 CHRF& 69.06 & 69.16 & 69.19 & 69.08& 69.16 &69.13&\textbf{69.68}\\
 COMET-Kiwi&82.30 &82.33 &\textbf{82.38} &82.36 &82.31 &82.37 &82.32\\
 \hline
\multicolumn{1}{c|}{} &\multicolumn{7}{c}{English-to-Chinese} \\ \hline
BLUE-4&25.82 &25.69 &25.87 &25.72 &25.63 &25.72 &\textbf{25.92} \\
METEOR  &56.65 &56.54 &56.64 &56.59 &56.52 &56.59 &\textbf{56.80} \\
ROUGE-1 &22.15 &22.21 &22.44 &22.35 &22.24 &22.16 &\textbf{22.47 } \\
ROUGE-2  &11.32 &11.21 &11.36 &11.24 &11.08 &11.23 &\textbf{11.39}  \\
ROUGE-L  &22.10 &22.14 &22.36 &22.28 &22.16 &22.08 &\textbf{22.40} \\
CHRF &48.46&48.33& 48.44&48.40&48.31&48.41&\textbf{49.78 } \\
COMET-Kiwi&88.98 &88.94 &\textbf{88.99} &88.97 &88.95 &\textbf{88.99} &\textbf{88.99}   \\
\hline
\end{tabular}
}

\caption{The translation performance of ChatGLM3-6B on the Diplomatic
Corpus dataset with different weights $\alpha_1$, $\alpha_2$, and $\alpha_3$. The best results are highlighted in \textbf{bold}.}
\label{tabel:7}
\end{table}

\subsubsection{Ablation on Weights of Hybrid Demonstration Retrieval} The hybrid demonstration retrieval uses multiple retrieval functions to compute relevance scores. To investigate the influence of different weights $\alpha_1$, $\alpha_2$, and $\alpha_3$, we conduct experiments with ChatGLM3-6B on the Diplomatic Corpus by setting different $\alpha_1$, $\alpha_2$, and $\alpha_3$. The results in Table~\ref{tabel:7} show that $\alpha_1$, $\alpha_2$, and $\alpha_3$ are set to 0.4, 0.4, and 0.2 can achieve the best performance on most of the metrics. 

\begin{table}[t]
\centering
\resizebox{\linewidth}{!}{
\begin{tabular}{c|ccc|ccc}
\hline
Demonstration number & 1        & 2        & 3           & 1        & 2        & 3          \\
\hline
\multicolumn{1}{c|}{}&\multicolumn{3}{c|}{Chinese-to-English} &\multicolumn{3}{c}{English-to-Chinese}\\
\hline
BLUE&26.71 &27.66 &\textbf{29.17} &24.24 &25.98 &\textbf{27.37} \\
METEOR&58.13 &58.91 &\textbf{60.02} &55.23 &56.91 &\textbf{58.23} \\
ROUGE-1&62.00 &63.63 &\textbf{64.02} &22.16 &22.04 &\textbf{22.92} \\
ROUGE-L&54.07 &54.72 &\textbf{55.15} &22.05 &21.95 &\textbf{22.85} \\
CHRF&68.71 &69.13 &\textbf{69.68} &47.23 &47.84&\textbf{49.78} \\
COMET-Kiwi&81.77 &82.34 &\textbf{82.74} &88.79 &88.92&\textbf{89.12}\\
\hline
\end{tabular}}

\caption{The translation performance of different numbers of demonstrations on the Diplomatic Corpus with ChatGLM3-6B. The parts highlighted in bold in the table indicate the best results.}
\label{table:8}
\end{table}

\subsubsection{The Number of Translation Demonstrations}
We verify the effects of different numbers of demonstrations for few-shot prompting by using the Diplomatic Corpus dataset on ChatGLM3-6B. The results in Table~\ref{table:8} show the best translation performance can be achieved when the number of demonstrations is 3.

\section{Conclusion}
In this work, we propose an adaptive few-shot prompting (AFSP) framework to automatically select suitable translation demonstrations for various source input sentences to further elicit the translation capability of an LLM for better machine translation. First, we retrieve top-k semantic-similar translation demonstrations from aligned parallel translation corpus based on hybrid demonstration retrieval. Then, to ensure better semantic consistency between source inputs and target outputs, we force the deployed LLM itself to generate multiple output candidates in the target language with the help of translation demonstrations and rerank these candidates. Besides, to better evaluate the effectiveness of our AFSP framework on the latest language and extend the research boundary of neural machine translation, we construct a high-quality diplomatic Chinese-English parallel dataset that consists of 5,528 parallel sentences. Extensive experiments on the proposed Diplomatic dataset and UN show the effectiveness and superiority of our AFSP.

\bibliography{aaai25}

\begin{thebibliography}{42}
\providecommand{\natexlab}[1]{#1}

\bibitem[{Agarwal et~al.(2024)Agarwal, Singh, Zhang, Bohnet, Chan, Anand,
  Abbas, Nova, Co-Reyes, Chu, Behbahani, Faust, and
  Larochelle}]{Agarwal2024ManyShotIL}
Agarwal, R.; Singh, A.; Zhang, L.~M.; Bohnet, B.; Chan, S.; Anand, A.; Abbas,
  Z.; Nova, A.; Co-Reyes, J.~D.; Chu, E.; Behbahani, F. M.~P.; Faust, A.; and
  Larochelle, H. 2024.
\newblock Many-Shot In-Context Learning.
\newblock \emph{ArXiv}, abs/2404.11018.

\bibitem[{Bahdanau, Cho, and Bengio(2014)}]{Bahdanau2014NeuralMT}
Bahdanau, D.; Cho, K.; and Bengio, Y. 2014.
\newblock Neural Machine Translation by Jointly Learning to Align and
  Translate.
\newblock \emph{CoRR}, abs/1409.0473.

\bibitem[{Bahdanau, Cho, and Bengio(2015)}]{bahdanau2015neural}
Bahdanau, D.; Cho, K.~H.; and Bengio, Y. 2015.
\newblock Neural machine translation by jointly learning to align and
  translate.
\newblock In \emph{3rd International Conference on Learning Representations,
  ICLR 2015}.

\bibitem[{Banerjee and Lavie(2005)}]{Banerjee2005METEORAA}
Banerjee, S.; and Lavie, A. 2005.
\newblock METEOR: An Automatic Metric for MT Evaluation with Improved
  Correlation with Human Judgments.
\newblock In \emph{IEEvaluation@ACL}.

\bibitem[{Bawden and Yvon(2023)}]{Bawden2023InvestigatingTT}
Bawden, R.; and Yvon, F. 2023.
\newblock Investigating the Translation Performance of a Large Multilingual
  Language Model: the Case of BLOOM.
\newblock In \emph{European Association for Machine Translation
  Conferences/Workshops}.

\bibitem[{Brown et~al.(2020)Brown, Mann, Ryder, Subbiah, Kaplan, Dhariwal,
  Neelakantan, Shyam, Sastry, Askell et~al.}]{brown2020language}
Brown, T.; Mann, B.; Ryder, N.; Subbiah, M.; Kaplan, J.~D.; Dhariwal, P.;
  Neelakantan, A.; Shyam, P.; Sastry, G.; Askell, A.; et~al. 2020.
\newblock Language models are few-shot learners.
\newblock \emph{Advances in neural information processing systems}, 33:
  1877--1901.

\bibitem[{Chen et~al.(2023)Chen, Zhang, Langren{\'e}, and
  Zhu}]{chen2023unleashing}
Chen, B.; Zhang, Z.; Langren{\'e}, N.; and Zhu, S. 2023.
\newblock Unleashing the potential of prompt engineering in Large Language
  Models: a comprehensive review.
\newblock \emph{arXiv preprint arXiv:2310.14735}.

\bibitem[{Chen et~al.(2024)Chen, Xiao, Zhang, Luo, Lian, and Liu}]{chen2024bge}
Chen, J.; Xiao, S.; Zhang, P.; Luo, K.; Lian, D.; and Liu, Z. 2024.
\newblock Bge m3-embedding: Multi-lingual, multi-functionality,
  multi-granularity text embeddings through self-knowledge distillation.
\newblock \emph{arXiv preprint arXiv:2402.03216}.

\bibitem[{Costa-juss{\`a} et~al.(2022)Costa-juss{\`a}, Cross, {\c{C}}elebi,
  Elbayad, Heafield, Heffernan, Kalbassi, Lam, Licht, Maillard
  et~al.}]{costa2022no}
Costa-juss{\`a}, M.~R.; Cross, J.; {\c{C}}elebi, O.; Elbayad, M.; Heafield, K.;
  Heffernan, K.; Kalbassi, E.; Lam, J.; Licht, D.; Maillard, J.; et~al. 2022.
\newblock No language left behind: Scaling human-centered machine translation.
\newblock \emph{arXiv preprint arXiv:2207.04672}.

\bibitem[{Devlin et~al.(2019)Devlin, Chang, Lee, and
  Toutanova}]{devlin-etal-2019-bert}
Devlin, J.; Chang, M.-W.; Lee, K.; and Toutanova, K. 2019.
\newblock {BERT}: Pre-training of Deep Bidirectional Transformers for Language
  Understanding.
\newblock In Burstein, J.; Doran, C.; and Solorio, T., eds., \emph{Proceedings
  of the 2019 Conference of the North {A}merican Chapter of the Association for
  Computational Linguistics: Human Language Technologies, Volume 1 (Long and
  Short Papers)}, 4171--4186. Minneapolis, Minnesota: Association for
  Computational Linguistics.

\bibitem[{Du et~al.(2022)Du, Qian, Liu, Ding, Qiu, Yang, and Tang}]{du2022glm}
Du, Z.; Qian, Y.; Liu, X.; Ding, M.; Qiu, J.; Yang, Z.; and Tang, J. 2022.
\newblock GLM: General Language Model Pretraining with Autoregressive Blank
  Infilling.
\newblock In \emph{Proceedings of the 60th Annual Meeting of the Association
  for Computational Linguistics (Volume 1: Long Papers)}, 320--335.

\bibitem[{Dubey et~al.(2024)Dubey, Jauhri, Pandey, Kadian, Al-Dahle, Letman,
  Mathur, Schelten, Yang, Fan et~al.}]{dubey2024llama}
Dubey, A.; Jauhri, A.; Pandey, A.; Kadian, A.; Al-Dahle, A.; Letman, A.;
  Mathur, A.; Schelten, A.; Yang, A.; Fan, A.; et~al. 2024.
\newblock The Llama 3 Herd of Models.
\newblock \emph{arXiv preprint arXiv:2407.21783}.

\bibitem[{Fan et~al.(2021)Fan, Bhosale, Schwenk, Ma, El-Kishky, Goyal, Baines,
  Celebi, Wenzek, Chaudhary et~al.}]{fan2021beyond}
Fan, A.; Bhosale, S.; Schwenk, H.; Ma, Z.; El-Kishky, A.; Goyal, S.; Baines,
  M.; Celebi, O.; Wenzek, G.; Chaudhary, V.; et~al. 2021.
\newblock Beyond english-centric multilingual machine translation.
\newblock \emph{Journal of Machine Learning Research}, 22(107): 1--48.

\bibitem[{Feng et~al.(2024{\natexlab{a}})Feng, Chen, Zhang, Meng, and
  Liu}]{feng2024ladder}
Feng, Z.; Chen, R.; Zhang, Y.; Meng, Z.; and Liu, Z. 2024{\natexlab{a}}.
\newblock Ladder: A Model-Agnostic Framework Boosting LLM-based Machine
  Translation to the Next Level.
\newblock \emph{arXiv preprint arXiv:2406.15741}.

\bibitem[{Feng et~al.(2024{\natexlab{b}})Feng, Zhang, Li, Wu, Liao, Liu, Lang,
  Feng, Wu, and Liu}]{Feng2024TEaRIL}
Feng, Z.; Zhang, Y.; Li, H.; Wu, B.; Liao, J.; Liu, W.; Lang, J.; Feng, Y.; Wu,
  J.; and Liu, Z. 2024{\natexlab{b}}.
\newblock TEaR: Improving LLM-based Machine Translation with Systematic
  Self-Refinement.

\bibitem[{Garc{\'i}a et~al.(2023)Garc{\'i}a, Bansal, Cherry, Foster, Krikun,
  Feng, Johnson, and Firat}]{Garca2023TheUE}
Garc{\'i}a, X.; Bansal, Y.; Cherry, C.; Foster, G.~F.; Krikun, M.; Feng, F.;
  Johnson, M.; and Firat, O. 2023.
\newblock The unreasonable effectiveness of few-shot learning for machine
  translation.
\newblock \emph{ArXiv}, abs/2302.01398.

\bibitem[{Guo et~al.(2024)Guo, Zhang, Ma, Zhang, and Feng}]{Guo2024SiLLMLL}
Guo, S.; Zhang, S.; Ma, Z.; Zhang, M.; and Feng, Y. 2024.
\newblock SiLLM: Large Language Models for Simultaneous Machine Translation.
\newblock \emph{ArXiv}, abs/2402.13036.

\bibitem[{Hendy et~al.(2023)Hendy, Abdelrehim, Sharaf, Raunak, Gabr,
  Matsushita, Kim, Afify, and Awadalla}]{Hendy2023HowGA}
Hendy, A.; Abdelrehim, M.~G.; Sharaf, A.; Raunak, V.; Gabr, M.; Matsushita, H.;
  Kim, Y.~J.; Afify, M.; and Awadalla, H.~H. 2023.
\newblock How Good Are GPT Models at Machine Translation? A Comprehensive
  Evaluation.
\newblock \emph{ArXiv}, abs/2302.09210.

\bibitem[{Jiang and Zhang(2024{\natexlab{a}})}]{jiang2024can}
Jiang, Z.; and Zhang, Z. 2024{\natexlab{a}}.
\newblock Can ChatGPT Rival Neural Machine Translation? A Comparative Study.
\newblock \emph{arXiv preprint arXiv:2401.05176}.

\bibitem[{Jiang and Zhang(2024{\natexlab{b}})}]{Jiang2024ConvergencesAD}
Jiang, Z.; and Zhang, Z. 2024{\natexlab{b}}.
\newblock Convergences and Divergences between Automatic Assessment and Human
  Evaluation: Insights from Comparing ChatGPT-Generated Translation and Neural
  Machine Translation.

\bibitem[{Jiao et~al.(2023)Jiao, Wang, tse Huang, Wang, and Tu}]{Jiao2023IsCA}
Jiao, W.; Wang, W.; tse Huang, J.; Wang, X.; and Tu, Z. 2023.
\newblock Is ChatGPT A Good Translator? A Preliminary Study.
\newblock \emph{ArXiv}, abs/2301.08745.

\bibitem[{Kirk et~al.(2023)Kirk, Mediratta, Nalmpantis, Luketina, Hambro,
  Grefenstette, and Raileanu}]{kirk2023understanding}
Kirk, R.; Mediratta, I.; Nalmpantis, C.; Luketina, J.; Hambro, E.;
  Grefenstette, E.; and Raileanu, R. 2023.
\newblock Understanding the effects of rlhf on llm generalisation and
  diversity.
\newblock \emph{arXiv preprint arXiv:2310.06452}.

\bibitem[{Lin(2004)}]{Lin2004ROUGEAP}
Lin, C.-Y. 2004.
\newblock ROUGE: A Package for Automatic Evaluation of Summaries.
\newblock In \emph{Annual Meeting of the Association for Computational
  Linguistics}.

\bibitem[{Merx et~al.(2024)Merx, Mahmudi, Langford, de~Araujo, and
  Vylomova}]{Merx2024LowResourceMT}
Merx, R.; Mahmudi, A.; Langford, K.; de~Araujo, L.~A.; and Vylomova, E. 2024.
\newblock Low-Resource Machine Translation through Retrieval-Augmented LLM
  Prompting: A Study on the Mambai Language.
\newblock \emph{ArXiv}, abs/2404.04809.

\bibitem[{Nori et~al.(2023)Nori, Lee, Zhang, Carignan, Edgar, Fusi, King,
  Larson, Li, Liu, Luo, McKinney, Ness, Poon, Qin, Usuyama, White, and
  Horvitz}]{nori2023generalistfoundationmodelsoutcompete}
Nori, H.; Lee, Y.~T.; Zhang, S.; Carignan, D.; Edgar, R.; Fusi, N.; King, N.;
  Larson, J.; Li, Y.; Liu, W.; Luo, R.; McKinney, S.~M.; Ness, R.~O.; Poon, H.;
  Qin, T.; Usuyama, N.; White, C.; and Horvitz, E. 2023.
\newblock Can Generalist Foundation Models Outcompete Special-Purpose Tuning?
  Case Study in Medicine.
\newblock arXiv:2311.16452.

\bibitem[{Ouyang et~al.(2022)Ouyang, Wu, Jiang, Almeida, Wainwright, Mishkin,
  Zhang, Agarwal, Slama, Ray et~al.}]{ouyang2022training}
Ouyang, L.; Wu, J.; Jiang, X.; Almeida, D.; Wainwright, C.; Mishkin, P.; Zhang,
  C.; Agarwal, S.; Slama, K.; Ray, A.; et~al. 2022.
\newblock Training language models to follow instructions with human feedback.
\newblock \emph{Advances in neural information processing systems}, 35:
  27730--27744.

\bibitem[{Papineni et~al.(2002)Papineni, Roukos, Ward, and
  Zhu}]{Papineni2002BleuAM}
Papineni, K.; Roukos, S.; Ward, T.; and Zhu, W.-J. 2002.
\newblock Bleu: a Method for Automatic Evaluation of Machine Translation.
\newblock In \emph{Annual Meeting of the Association for Computational
  Linguistics}.

\bibitem[{Popovi{\'c}(2015)}]{popovic-2015-chrf}
Popovi{\'c}, M. 2015.
\newblock chr{F}: character n-gram {F}-score for automatic {MT} evaluation.
\newblock In Bojar, O.; Chatterjee, R.; Federmann, C.; Haddow, B.; Hokamp, C.;
  Huck, M.; Logacheva, V.; and Pecina, P., eds., \emph{Proceedings of the Tenth
  Workshop on Statistical Machine Translation}, 392--395. Lisbon, Portugal:
  Association for Computational Linguistics.

\bibitem[{Rei et~al.(2022)Rei, Treviso, Guerreiro, Zerva, Farinha, Maroti,
  C.~de Souza, Glushkova, Alves, Coheur, Lavie, and Martins}]{comet2022}
Rei, R.; Treviso, M.; Guerreiro, N.~M.; Zerva, C.; Farinha, A.~C.; Maroti, C.;
  C.~de Souza, J.~G.; Glushkova, T.; Alves, D.; Coheur, L.; Lavie, A.; and
  Martins, A. F.~T. 2022.
\newblock {C}omet{K}iwi: {IST}-Unbabel 2022 Submission for the Quality
  Estimation Shared Task.
\newblock In Koehn, P.; Barrault, L.; Bojar, O.; Bougares, F.; Chatterjee, R.;
  Costa-juss{\`a}, M.~R.; Federmann, C.; Fishel, M.; Fraser, A.; Freitag, M.;
  Graham, Y.; Grundkiewicz, R.; Guzman, P.; Haddow, B.; Huck, M.; Jimeno~Yepes,
  A.; Kocmi, T.; Martins, A.; Morishita, M.; Monz, C.; Nagata, M.; Nakazawa,
  T.; Negri, M.; N{\'e}v{\'e}ol, A.; Neves, M.; Popel, M.; Turchi, M.; and
  Zampieri, M., eds., \emph{Proceedings of the Seventh Conference on Machine
  Translation (WMT)}, 634--645. Abu Dhabi, United Arab Emirates (Hybrid):
  Association for Computational Linguistics.

\bibitem[{Sennrich, Haddow, and Birch(2015)}]{Sennrich2015NeuralMT}
Sennrich, R.; Haddow, B.; and Birch, A. 2015.
\newblock Neural Machine Translation of Rare Words with Subword Units.
\newblock \emph{ArXiv}, abs/1508.07909.

\bibitem[{Touvron et~al.(2023)Touvron, Martin, Stone, Albert, Almahairi,
  Babaei, Bashlykov, Batra, Bhargava, Bhosale et~al.}]{touvron2023llama}
Touvron, H.; Martin, L.; Stone, K.; Albert, P.; Almahairi, A.; Babaei, Y.;
  Bashlykov, N.; Batra, S.; Bhargava, P.; Bhosale, S.; et~al. 2023.
\newblock Llama 2: Open foundation and fine-tuned chat models.
\newblock \emph{arXiv preprint arXiv:2307.09288}.

\bibitem[{Vilar et~al.(2022)Vilar, Freitag, Cherry, Luo, Ratnakar, and
  Foster}]{Vilar2022PromptingPF}
Vilar, D.; Freitag, M.; Cherry, C.; Luo, J.; Ratnakar, V.; and Foster, G.~F.
  2022.
\newblock Prompting PaLM for Translation: Assessing Strategies and Performance.
\newblock \emph{ArXiv}, abs/2211.09102.

\bibitem[{Wang et~al.(2023)Wang, Lyu, Ji, Zhang, Yu, Shi, and
  Tu}]{Wang2023DocumentLevelMT}
Wang, L.; Lyu, C.; Ji, T.; Zhang, Z.; Yu, D.; Shi, S.; and Tu, Z. 2023.
\newblock Document-Level Machine Translation with Large Language Models.
\newblock In \emph{Conference on Empirical Methods in Natural Language
  Processing}.

\bibitem[{Wang et~al.(2022)Wang, Jiao, Hao, Wang, Shi, Tu, and
  Lyu}]{Wang2022UnderstandingAI}
Wang, W.; Jiao, W.; Hao, Y.; Wang, X.; Shi, S.; Tu, Z.; and Lyu, M.~R. 2022.
\newblock Understanding and Improving Sequence-to-Sequence Pretraining for
  Neural Machine Translation.
\newblock In \emph{Annual Meeting of the Association for Computational
  Linguistics}.

\bibitem[{Wu et~al.(2024)Wu, Yuan, Haffari, and Wang}]{Wu2024PerhapsBH}
Wu, M.; Yuan, Y.; Haffari, G.; and Wang, L. 2024.
\newblock (Perhaps) Beyond Human Translation: Harnessing Multi-Agent
  Collaboration for Translating Ultra-Long Literary Texts.
\newblock \emph{ArXiv}, abs/2405.11804.

\bibitem[{Yao et~al.(2023)Yao, Chen, Zou, Lu, Li, Zhang, Liu, Hendler, and
  Wang}]{yao2023more}
Yao, B.; Chen, G.; Zou, R.; Lu, Y.; Li, J.; Zhang, S.; Liu, S.; Hendler, J.;
  and Wang, D. 2023.
\newblock More Samples or More Prompt Inputs? Exploring Effective In-Context
  Sampling for LLM Few-Shot Prompt Engineering.
\newblock \emph{arXiv preprint arXiv:2311.09782}.

\bibitem[{Yuan et~al.(2023)Yuan, Lu, Zhu, Kong, Li, Qiao, and
  Xu}]{yuan-etal-2023-lego}
Yuan, F.; Lu, Y.; Zhu, W.; Kong, L.; Li, L.; Qiao, Y.; and Xu, J. 2023.
\newblock {L}ego-{MT}: Learning Detachable Models for Massively Multilingual
  Machine Translation.
\newblock In Rogers, A.; Boyd-Graber, J.; and Okazaki, N., eds., \emph{Findings
  of the Association for Computational Linguistics: ACL 2023}, 11518--11533.
  Toronto, Canada: Association for Computational Linguistics.

\bibitem[{Zeng et~al.(2022)Zeng, Liu, Du, Wang, Lai, Ding, Yang, Xu, Zheng, Xia
  et~al.}]{zeng2022glm}
Zeng, A.; Liu, X.; Du, Z.; Wang, Z.; Lai, H.; Ding, M.; Yang, Z.; Xu, Y.;
  Zheng, W.; Xia, X.; et~al. 2022.
\newblock Glm-130b: An open bilingual pre-trained model.
\newblock \emph{arXiv preprint arXiv:2210.02414}.

\bibitem[{Zhang, Haddow, and Birch(2023{\natexlab{a}})}]{zhang2023prompting}
Zhang, B.; Haddow, B.; and Birch, A. 2023{\natexlab{a}}.
\newblock Prompting large language model for machine translation: A case study.
\newblock In \emph{International Conference on Machine Learning}, 41092--41110.
  PMLR.

\bibitem[{Zhang, Haddow, and Birch(2023{\natexlab{b}})}]{Zhang2023PromptingLL}
Zhang, B.; Haddow, B.; and Birch, A. 2023{\natexlab{b}}.
\newblock Prompting Large Language Model for Machine Translation: A Case Study.
\newblock \emph{ArXiv}, abs/2301.07069.

\bibitem[{Zhang et~al.(2023)Zhang, Rajabi, Duh, and
  Koehn}]{zhang-etal-2023-machine}
Zhang, X.; Rajabi, N.; Duh, K.; and Koehn, P. 2023.
\newblock Machine Translation with Large Language Models: Prompting, Few-shot
  Learning, and Fine-tuning with {QL}o{RA}.
\newblock In Koehn, P.; Haddow, B.; Kocmi, T.; and Monz, C., eds.,
  \emph{Proceedings of the Eighth Conference on Machine Translation}, 468--481.
  Singapore: Association for Computational Linguistics.

\bibitem[{Zhu et~al.(2024)Zhu, Liu, Dong, Xu, Huang, Kong, Chen, and
  Li}]{zhu2024multilingual}
Zhu, W.; Liu, H.; Dong, Q.; Xu, J.; Huang, S.; Kong, L.; Chen, J.; and Li, L.
  2024.
\newblock Multilingual Machine Translation with Large Language Models:
  Empirical Results and Analysis.
\newblock In \emph{Findings of the Association for Computational Linguistics:
  NAACL 2024}, 2765--2781.

\end{thebibliography}

\end{document}